\pdfoutput=1

\documentclass[11pt]{article}

\usepackage{acl}
\usepackage{times}
\usepackage{latexsym}
\usepackage{soul}

\usepackage[T1]{fontenc}

\usepackage[utf8]{inputenc}

\usepackage{microtype}
\usepackage{booktabs} 
\usepackage{enumitem} 
\usepackage{graphicx}
\usepackage{multicol}
\usepackage{multirow}
\usepackage{makecell}
\usepackage{amsmath}
\usepackage{amssymb}
\usepackage{amsfonts}
\usepackage{bbm}
\usepackage[normalem]{ulem} 

\newcommand*\samethanks[1][\value{footnote}]{\footnotemark[#1]}

\makeatletter
\newcommand{\printfnsymbol}[1]{%
  \textsuperscript{\@fnsymbol{#1}}%
}
\DeclareMathOperator*{\E}{\mathbb{E}}
\DeclareMathOperator{\R}{\mathbb{R}}
\DeclareMathOperator{\x}{\mathbf{x}}
\DeclareMathOperator{\y}{\mathbf{y}}
\DeclareMathOperator*{\minimize}{minimize}

\usepackage{color, colortbl}
\definecolor{lightgreen}{rgb}{0.91,1,0.91}
\definecolor{lightred}{rgb}{1.0, 0.91, 0.91}
\title{\textit{Generalized but not Robust?} Comparing the Effects of Data Modification Methods on Out-of-Domain Generalization and Adversarial Robustness}

\author{
    Tejas Gokhale\thanks{~~Equal Contribution}
    \qquad  Swaroop Mishra\samethanks
    \qquad Man Luo\samethanks \\
    {\bf  Bhavdeep Singh Sachdeva
    \qquad Chitta Baral} \\
    Arizona State University \\
    \texttt{\{tgokhale, srmishr1, mluo26, bssachde, chitta\}@asu.edu}
}

\begin{document}
\maketitle

\begin{abstract}
    Data modification, either via additional training datasets, data augmentation, debiasing, and dataset filtering, has been proposed as an effective solution for generalizing to out-of-domain (OOD) inputs, in both natural language processing and computer vision literature.
    However, the effect of data modification on adversarial robustness remains unclear.
    In this work, we conduct a comprehensive study of common data modification strategies and evaluate not only their in-domain and OOD performance, but also their adversarial robustness (AR).
    We also present results on a two-dimensional synthetic dataset to visualize 
    the effect of each method on the training distribution.
    This work serves as an empirical study towards understanding the relationship between generalizing to unseen domains and defending against adversarial perturbations.
    Our findings suggest that more data (either via additional datasets or data augmentation) benefits both OOD accuracy and AR.
    However, data filtering (previously shown to improve OOD accuracy on natural language inference) hurts OOD accuracy on other tasks such as question answering and image classification.
    We provide insights from our experiments to inform future work in this direction.
\end{abstract}

\section{Introduction}
Deep neural networks have emerged as a widely popular architectural choice for modeling tasks in multiple domains such as (but not limited to) computer vision~\cite{yuille2021deep}, natural language processing~\cite{hochreiter1997long,vaswani2017attention}, and audio~\cite{hannun2014deep}.
While these models are highly capable of learning from training data, recent studies show that they are quite prone to failure on new test sets or under distribution shift~\cite{taori2020measuring}, natural corruptions~\cite{hendrycks2018benchmarking}, adversarial attacks~\cite{goodfellow2014explaining}, spurious correlations~\cite{beery2018recognition}, and many other types of ``unseen'' changes that may be encountered after training.
This shortcoming stems from the \textit{i.i.d.} assumption in statistical machine learning which guarantees good performance only on test samples that are drawn from an underlying distribution that is identical to the training dataset.
For instance, digit recognition models trained on the black-and-white MNIST training images are almost perfect ($>99\%$ accuracy) on the corresponding test set, yet their performance on colored digits and real-world digits from street number plates is less than $75\%$.
Similarly, state-of-the-art NLP models have been shown to fail when negation is introduced in the input~\cite{kassner2020negated}.
These findings pose a significant challenge to the practical adoption of these models and their reliability in the real-world.

To test model performance beyond the traditional notion of in-domain (ID) generalization, two prominent ideas have emerged: 
out-of-domain (OOD generalization) \textit{a.k.a.} domain generalization\footnote{In this paper we use these two terms interchangeably.}, and adversarial robustness.
The OOD generalization objective expects a model which is trained on distribution $\mathcal{D}$ to perform reliably on unseen distributions $\mathcal{D}_{e}, e\in\{1,\dots,n\}$, that differ from $\mathcal{D}$.
For a trained classifier $f^*$, OOD accuracy on previously unseen distribution $\mathcal{D}_e$ is defined as:
\begin{equation}
    \textrm{acc}_{\textrm{OOD}}^e ={\E_{(\x,\y) \sim \mathcal{D}_e}} [\mathbb{I}(f^*(\x) = \y)] 
\end{equation}
To define adversarial robustness, consider an input $\x$ and a true label $\y$.
For a classifier loss function $\ell$, a loss-maximizing perturbation $\delta^*$ within $\Delta_\epsilon$ (an $\epsilon$-bounded neighborhood of $\x$) is defined as: 
\begin{equation}
    \delta^{*}_{\x} = \underset{\delta\in\Delta_\epsilon}{\textrm{max}}~~\ell(f^*(\x+\delta), \y).
\end{equation}

The second idea is that of adversarial robustness.
Recent work on adversarial examples has revealed the vulnerability of deep neural networks against small perturbations of the original data.
Adversarial robustness in such under this setting is defined as the accuracy of the classifier on adversarial samples $\x+\delta_{\x}$, where the perturbation lies within an $\ell_p$ norm bound: $||\delta_{\x}||_p < \epsilon$.
\begin{equation}
    \textrm{acc}_{rob} = \E_{(\x,\y)\sim\mathcal{D}}\mathbb{I}(f^*(\x+\delta_{\x})=\y).
\end{equation}
In the context of text classification, the norm-bound can also be in the form of small character-level or word-level perturbations such as swapping, inserting, or deleting characters or words.
In essence, adversarial robustness measures the invariance of the classifier to small perturbations of the input.

Various methods have been developed that either improve OOD generalization or improve adversarial robustness.
Notable among these are techniques that modify the distribution of the training dataset.
In this paper, we focus on three major data modification techniques -- the use of additional datasets (also known as multi-source training), data augmentation, and data filtering; in addition we also consider model-based debiasing techniques which do not alter the data distribution explicitly.
We study the performance of these methods on three representative tasks -- natural language inference (NLI), extractive question answering (QA), and image classification (IC).

Our first aim in this paper is to understand whether the increase or decrease in OOD generalization by each method over the naive baseline (standard training on the source dataset) is consistent across tasks.
To further conduct fine-grained analysis, we also analyze the effect of these methods on in-domain (ID) accuracy on the test set for each task, since in the ideal case improvement in OOD performance should not come at the cost of in-domain accuracy.

Recent work seeks to understand the relationships between in-domain and out-of-domain performance: for instance,~\citet{miller2021accuracy} empirically show that ID and OOD performance are strongly correlated,~\citet{raghunathan2020understanding,yang2020closer} show a trade-off between robustness and accuracy for adversarially trained models.
However it is not clear how methods \textit{designed for OOD generalization} affect robustness.
This is largely because work on domain generalization reports only IID and OOD metrics, and work on robustness reports only ID and robustness metrics.
Our second aim is to understand the effect of these generalization methods on adversarial robustness.

In addition to our experiments on NLP and vision tasks, we also provide an experiment on a synthetic binary classification dataset where points lie in a 2-dimensional feature space and are separated by concentric circles into class labels.
This setting allows us to visualize the effect of data modification techniques on the training distribution and the resulting performance.\\

\noindent Our findings can be summarized as follows:
\begin{itemize}[noitemsep]
    \item More data benefits OOD generalization, 
    \item Data filtering hurts OOD generalization, and
    \item Data filtering significantly hurts adversarial robustness on all benchmarks.
\end{itemize}
These findings and our additional analysis raise new questions for robustness and domain generalization research.
Significant among these are the importance of both diversity and number of training samples for inductive bias and generalization guarantees, the problems associated with data filtering in terms of robustness, and the importance of a comprehensive set of evaluation metrics that could be adopted for future work.

\section{Categorization of Domain Generalization Methods} 
\label{sec:categ}
In this section, we provide a categorization of methods that are typically used as baselines for domain generalization.
We briefly explain the method and provide relevant related work in which these ideas are used as methods for domain generalization.
Throughout this paper, we will refer to the original training distribution as the \textit{``source''} and the out-of-distribution datasets as the \textit{``targets''}.

\paragraph{Single-Source Training}(\texttt{SS}) refers to the ``vanilla'' baseline which is trained only on the source dataset, without any dataset modification.
\texttt{SS} utilizes no other information apart from the single source dataset $\mathcal{D}$ and updates parameters $\theta$ of classifier $f$ to minimize the risk on the source using approaches such as ERM~\citep{vapnik1991necessary}. 
\begin{equation}
    \minimize_{\theta} \E_{(\x, \y) \sim \mathcal{D}}  ~\ell(f(\x; \theta), \y).
    \label{eq:ss_erm}
\end{equation}

\paragraph{Multi-Source Training}(\texttt{MS}).
This method is identical to \texttt{SS} except that additional training datasets $\mathcal{D}^\prime$ are used for risk minimization.
\begin{equation}
    \minimize_{\theta} \E_{(\x, \y) \sim \mathcal{D}\cup \mathcal{D}^\prime}  ~\ell(f(\x; \theta), \y).
    \label{eq:ms_erm}
\end{equation}
Usually $\mathcal{D}^\prime$ are designed for the same task as  $\mathcal{D}$ but may have different styles, characteristics, or sources of collection.
For instance, while both SNLI~\citep{bowman2015large} and MNLI~\cite{williams2018broad} are datasets for natural language inference with identical class labels, SNLI was collected from image captions, while MNLI was collected from Open American National Corpus\footnote{\url{https://www.anc.org/}}.

\citet{gulrajani2020search} provide an extensive comparitive study of models trained for multi-source domain generalization for image classification and surprisingly find that if multiple source domains are available, ERM is empirically the best approach as compared to specially designed DG methods such as meta-learning~\cite{li2018learning}, learning domain-invariant features~\cite{ganin2016domain}, invariant risk minimization~\cite{arjovsky2019invariant}, etc.
These findings have also been observed on text classification experiments in~\cite{koh2021wilds}.
\citet{hendrycks2020pretrained} show that pre-training transformer architectures on diverse data leads to higher OOD accuracies on multiple tasks such as semantic textual similarity, sentiment classification, reading comprehension and natural language inference.

\paragraph{Data Augmentation}(\texttt{DA}).
When additional training distributions are not directly available, transformations of samples in $\mathcal{D}$ using pre-defined augmentation functions can be used to create $\mathcal{D}^\prime$ and train the model.
Such data augmentation functions are typically derived from existing knowledge about the invariance of the task w.r.t.\ certain transformations. 
For instance, for image classification, addition of small noise, small translations, scaling, etc. are common data augmentation functions, since they do not change the true label for the image.
Similarly, for text inputs, synonyms of words are commonly used since they do not change the semantics of the sentence.
NLP data augmentation techniques include UDA~\cite{xie2020unsupervised}, EDA~\cite{wei2019eda}, and back-translation for question answering~\citep{longpre-etal-2019-exploration}.

\paragraph{Data Filtering}(\texttt{DF}).
Dataset filtering has been previously explored for quality control, such as, removing noise and artifacts to curate and improve publicly sourced datasets.
However, there has been recent interest in considering \texttt{DF} as a method for bias reduction and generalization.
This idea can be traced back to work by ~\citet{zellers2018swag,zellers2019recognition}, that proposed \texttt{DF} as an algorithmic method to avoid annotation artifacts and spurious correlations during dataset construction.
AFLite~\cite{le2020adversarial} extended this idea to a generic filtering methodology that can work without any pre-defined rules or strategies.
Instead, AFLite operates by utilizing several weak learners (such as support-vector machines) trained over small subsets to identify samples that are easy to classify.
It is argued that such samples are more likely to carry biases, and as such, could be removed.
AFLite suggests that reduction of a dataset to even $10\%$ of the original size can boost OOD accuracy on NLI.
In the vision domain, similar ideas have been proposed concurrently, including REPAIR~\cite{li2019repair} and RESOUND~\cite{li2018resound}, in which instead of completely removing samples, biased samples are assigned smaller weights.
However these methods require a prior knowledge of the bias variable.
\citet{liu2021just} have recently proposed a simple approach which upweights samples which have higher loss -- this is shown to improve worst-group accuracy without having access to the bias variable.

\paragraph{Model De-biasing}(\texttt{DB}).
Methods under this category do not directly alter the training dataset, but instead resort to changes in the modeling technique -- these changes can be in terms of the optimization function, regularization, additional auxiliary costs, etc.
The main idea in \texttt{DB} is to utilize known biases (or identify unknown biases) in the data distribution, model these biases in the training pipeline, and use this knowledge to train robust classifiers~\cite{clark2019don,wu2020improving,bhargava-etal-2021-generalization}.
In the image classification literature, there is growing consensus on enforcing a consistency on different views (or augmentations) of an image in order to achieve debiasing~\cite{hendrycks2019augmix,xu2020robust,chai2021ensembling,nam2021reducing}.
Unlike \texttt{DF}, model de-biasing does not directly alter the training distribution, but instead allows the model to learn which biases to ignore.

\section{Toy Example: Concentric Circles}

\begin{figure}[t]
    \centering
    \includegraphics[width=0.9\linewidth]{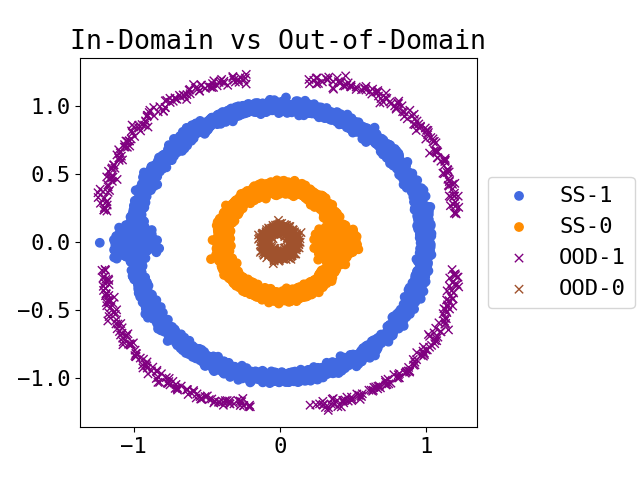}
    \caption{Our toy experimental setting consists of points in $\R^2$ belonging to two classes (0/1).
    This illustration shows the discrepancy between the source dataset (\texttt{SS}) and the out-of-domain dataset (\texttt{OOD}).
    }
    \label{fig:viz_iid_ood}
\end{figure}

\begin{figure*}
    \centering
    \includegraphics[width=\linewidth]{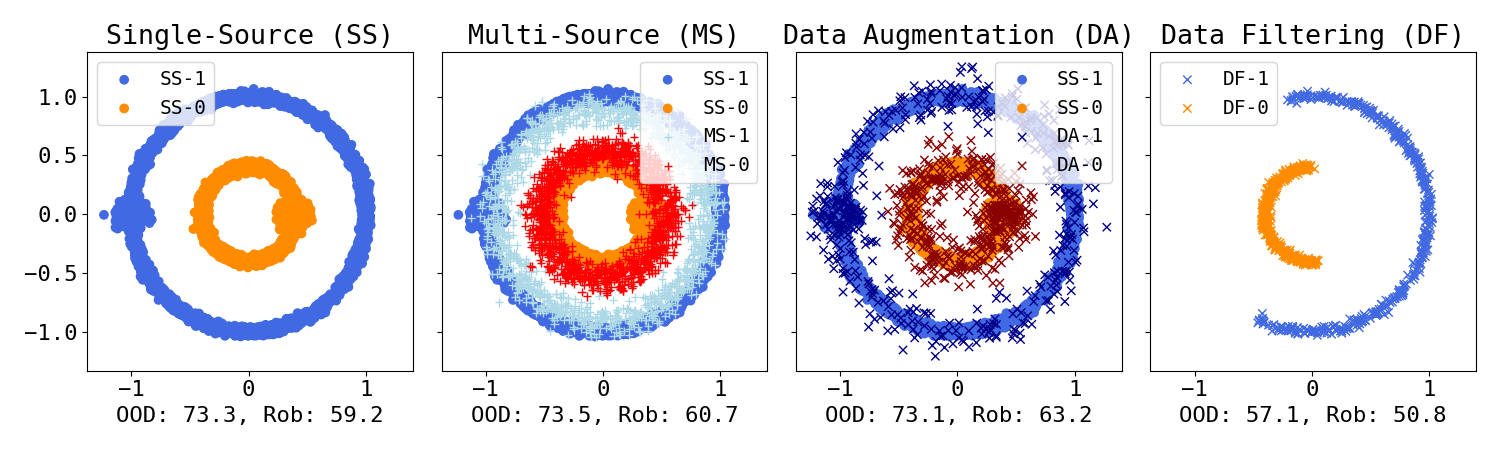}
    \caption{This figure illustrates the effect of data modification techniques on the training distribution.
    The leftmost figure shows the training distribution in the single-source setting.
    The introduction of a second dataset or Data-augmentation (done using small perturbations of source samples with Gaussian noise) makes the distribution more diverse in the multi-source (\texttt{MS}) and data augmentation (\texttt{DA}) setting respectively.
    On the other hand, data filtering, in order to remove spurious correlations from the dataset, removes points from certain sectors of the distribution.
    The effect of each strategy on OOD generalization and robustness is shown below each plot.
    }
    \label{fig:viz_effect_on_distribution}
\end{figure*}

We begin with a simple two-dimensional example to illustrate our experimental setting and to show how each method affects the distribution of the training set.
Consider the set of points shown in Figure~\ref{fig:viz_iid_ood} where the points belong to two class labels (either \texttt{0} or \texttt{1}) and are seen to lie on concentric circles.
Points with label \texttt{0} are closer to the origin, while points with label \texttt{1} are closer to a distance of \texttt{1} from the origin.
Our aim is to start with the single source dataset and train the model to generalize on the out-of-domain (OOD) dataset.
An important thing to note here is that the source dataset contains a subset of points with label \texttt{0} (orange) clustered around $(0.4, 0.0)$ and a subset with label \texttt{1} clustered around $(-1, 0.0)$.
This implies that class-\texttt{0} is biased towards $x>0$, while class-\texttt{1} is biased towards $x<0$.
In total, our \texttt{SS} dataset consists of $10000$ samples, of which $20\%$ are biased.

We apply three data modifications: additional source (\texttt{MS}), gaussian data augmentation (\texttt{DA}) ${\sim}\mathcal{N}(0, 0.1)$, and data filtering (AFLite) which reduces the dataset size to $10\%$.
Note that we do not show model debiasing (\texttt{DB}) here, since it does not alter the data distribution.
Figure~\ref{fig:viz_effect_on_distribution} shows the effect on the data distribution.
The most striking is the effect of \texttt{DF} which removes all samples previously in the biased clusters near $(0.4, 0.0)$ and $(-1.0, 0.0)$.

Equipped with these resulting datasets, we train a linear SGD classifier with log-loss and evaluate the robustness of each model in terms of in-domain and OOD accuracies.
We also evaluate adversarial robustness by using standard PGD attacks.
Results are shown in the textboxes in Figure~\ref{fig:viz_effect_on_distribution}.
It can be seen that data filtering significantly hurts both OOD generalization and robustness.
This finding motivates our experiments to understand the effect of each method for NLP and vision tasks.

\begin{table*}[t]
    \centering 
    \small
    \resizebox{\linewidth}{!}{
    \begin{tabular}{@{}llll@{}}
        \toprule
        \multirow{2}{*}{\textbf{Method Category}} &  \multicolumn{3}{c}{\textbf{ Tasks}} \\
        \cmidrule{2-4}
         &  {Natural Language Inference} & {Question Answering} & {Image Classification} \\
        \midrule 
        \texttt{SS} (Single-Source ERM)    &  {SNLI} & NQ~\citep{kwiatkowski2019natural} & MNIST\\
        \texttt{MS} (Multi-Source ERM)     &  {SNLI~+~MNLI} & NQ~+~SQuAD+NQA+HQA+SQA+TQA & MNIST~+~USPS\\
        \texttt{DA} (Data Augmentation)    &  EDA~\cite{wei2019eda} & {QG~\citep{chan2019recurrent}} & M-ADA~\cite{qiao2020learning}\\
        \texttt{DB} (Model De-biasing)     &  {LMH~\citep{clark2019don}} & Mb-CR\citep{wu2020improving} & RandConv~\cite{xu2020robust}\\
        \texttt{DF} (Data Filtering)       & AFLite~\cite{le2020adversarial} & AFLite (adapted for QA) & AFLite \\
        \bottomrule
    \end{tabular}
    }
    \caption{
    List of method categories and specific methods that we use under each task setting in nour experiments.
    Details for each can be found in Section~\ref{sec:exp} for the corresponding task.
    }
    \label{tab:methods}
\end{table*}

\section{Experiments}
\label{sec:exp}
In this section, we present three tasks and their corresponding experimental setup, evaluation protocol and our findings. 
A summary of methods belong to each category is provided in Table~\ref{tab:methods} and the abbreviations \texttt{SS}, \texttt{MS}, \texttt{DA}, \texttt{DB}, \texttt{DF} are used henceforth.

\subsection{Natural Language Inference (NLI)}
\begin{table*}[t]
    \centering 
    \small
    \resizebox{\linewidth}{!}{
    \begin{tabular}{@{}lc ccc c ccc c ccccc@{}}
        \toprule
        \multirow{3}{*}{\textbf{Method}} & \multirow{3}{*}{\textbf{\makecell{In-Domain \\ Acc. (\%)}}} & \multicolumn{12}{c}{\textbf{OOD Acc. (\%)}} \\
         & & \multicolumn{4}{c}{NLI-Diagnostics} & \hphantom & \multicolumn{3}{c}{Stress Test} & \hphantom & \multicolumn{3}{c}{HANS} & \multirow{3}{*}{\textbf{Avg}} \\
         \cmidrule{3-6} \cmidrule{8-10} \cmidrule{12-14}
         & & {Kno.} & {Lex.} & {Log.} & {PAS} && {Comp.} & {Distr.} & {Noise} && {Lex.} & {Subs.} & {Consti.} &  \\
        \midrule 
        \texttt{{SS}} & \textbf{89.6} & {51.8} & {65.7} & {57.8} &{72.6} && {77.9} & {73.5} & {79.8} && {88.4} & {28.2} & {21.7} & 61.74 \\
        \texttt{{MS}} & \cellcolor{lightred}{87.8} & {52.1} & {66.8} & {57.8} &{72.8} && {79.6} & {72.4} & {79.2} && {92.0} & {33.6} & {26.7} & \cellcolor{lightgreen}{63.30} \\
        \texttt{{DA}} & \cellcolor{lightred}{87.2} & {52.1} & {66.0} & {58.1} &{72.6} && {79.6} & {71.8} & {79.2} && {92.8} & {32.8} & {26.4} & \cellcolor{lightgreen}{{63.14}} \\
        \texttt{{DB}} & \cellcolor{lightred}{81.8} & {52.4} & {66.0} & {58.4} &{72.8} && {79.3} & {71.8} & {79.5} && {92.2} & {33.8} & {27.5} & \cellcolor{lightgreen}{63.37} \\
        \texttt{{DF}} & \cellcolor{lightred}{62.6} & {53.9} & {66.5} & {58.7} &{68.9} && {79.1} & {72.0} & {79.5} && {94.1} & {46.3} & {38.5} & \cellcolor{lightgreen}{\bf 65.75} \\
        \bottomrule
    \end{tabular}
    }
    \caption{
    NLI Result: In-domain (IID) accuracy and out-of-domain generalization (OOD) on the NLI benchmark using SNLI as source dataset.
    \footnote{Green$\rightarrow$increase and Red$\rightarrow$decrease w.r.t.~\texttt{SS}.}
    \textit{See Table~\ref{tab:methods} for method abbreviations.}  
    }
    \label{tab:nli_iid_ood}
\end{table*}

\begin{table*}[t]
    \centering 
    \small
    \begin{tabular}{@{}lcccccccc@{}}
        \toprule
        \multirow{2}{*}{\textbf{Method}} & \multirow{2}{*}{\textbf{\makecell{Model Based \\ \#Num Queries}}} & \multicolumn{7}{c}{\textbf{Model Free Accuracy (\%)}} \\
        \cmidrule{3-9}
         & & {CharSwap} & {EasyData} & {Embedding} &{WordNet} & {CheckList} & {CLARE} & \textbf{Avg} \\
        \midrule 
        \texttt{{SS}} & 53.56 & 81.3 & 72.0 & 81.9 &77.0 & 89.4 & 76.3 & 79.65\\
        \texttt{{MS}} & \cellcolor{lightgreen}{54.44} & 81.5 & 71.6 & 82.0 &78.2 & 89.2 & 77.5 & \cellcolor{lightgreen}{80.00} \\
        \texttt{{DA}} & \cellcolor{lightgreen}{\bf 55.06} & 77.7 & 74.1 & 80.7 &80.2 & 86.6 & 80.5 & \cellcolor{lightgreen}{79.97} \\
        \texttt{{DB}} & \cellcolor{lightgreen}{54.82} & 81.5 & 72.4 & 82.3 &78.0 & 89.2 & 77.0 & \cellcolor{lightgreen}{\bf 80.07} \\
        \texttt{{DF}} & \cellcolor{lightred}{51.13} & 65.2 & 56.8 & 66.2 & 62.5 & 72.3 & 62.5 & \cellcolor{lightred}{64.25} \\
        \bottomrule
    \end{tabular}
    \caption{
    NLI Result: Comparison of robustness in terms of model-based evaluation (number of queries needed to fool the model) and model-free (accuracy on adversarial transformations).
    \footnotemark[2]
    \textit{See Table~\ref{tab:methods} for method abbreviations.}
    }
    \label{tab:nli_robustness}
\end{table*}

NLI is the task of determining whether a \textit{hypothesis} is true (entailment), false (contradiction), or undetermined (neutral) given a \textit{premise}.

\paragraph{Methods.}
We use RoBERTa as the backbone model for each method and SNLI~\citep{bowman2015large} as our source training corpus.
A model trained with expected risk minimization (ERM) on SNLI alone, forms our single-source (\texttt{SS}) baseline.
A model trained with a combination of SNLI and MNLI~\citep{williams2018broad} forms our multi-source (\texttt{MS}) baseline.
We apply EDA~\citep{wei2019eda} to augment our training dataset with $100\%$ of additional data to train a \texttt{DA} model. 
The LMH debiasing method from~\citet{clark2019don} represents our \texttt{DB} model.
For data filtering, we use AFlite~\citep{le2020adversarial} to filter out $90\%$ of the SNLI training data, and use the remaining $10\%$ data to train our \texttt{DF} model -- this setting is based on the experiments from~\cite{le2020adversarial}.

\paragraph{Evaluation Protocol.}
We report accuracy on the SNLI test set (IID), and to evaluate generalization, we report accuracy on NLI diagnostics ~\citep{wang2018glue}, Stress test evaluation~\citep{naik2018stress} and HANS~\citep{mccoy2019right}.
We use two metrics for evaluating robustness:
\begin{itemize}[nosep,noitemsep,leftmargin=*]
    \item \textit{model-based robustness} uses BAE adversarial attack~\citep{garg2020bae}, implemented using TextAttack~\citep{morris2020textattack}, and reports robustness as number of queries (sequential perturbations) needed to fool the model. 
    \item \textit{model-free robustness} uses six pre-defined operations to transform SNLI test inputs into adversarial examples. These six methods are: CLARE~\cite{li2021contextualized}, character-swap~\cite{pruthi2019combating}, Checklist~\cite{ribeiro2020beyond}, EDA~\cite{wei2019eda}, counter-fitted embeddings (Emb)~\cite{alzantot2018generating}.
\end{itemize}

\paragraph{Results.}
Table~\ref{tab:nli_iid_ood} shows the performance of each method in terms of in-domain and out-of-domain accuracy.
We observe that four methods all improve the generalization performance on average but decrease the in-domain performance.
Especially, \texttt{DF} method is the best in terms of OOD accuracy, but is the worst in terms of in-domain performance.
We also see a trend that four methods improve the generalization in all sets of NLI-Diagnostics and HANS, while all four methods do not show improvement on  generalization on Distraction and Noise sets of Stress dataset. 

Table~\ref{tab:nli_robustness} shows the robustness evaluation. 
We see that except for \texttt{DF}, all methods improve the robustness under both model-based and model-free evaluation. 
\texttt{MS} improves the robustness in all transformations except for EDA. 
\texttt{DA} achieves the best robustness by model-based evaluation but is not consistent in terms of different transformations of model-free evaluation. 
\texttt{DB} improves the robustness in terms of every transformation and achieves the best robustness in terms of average of model-free evaluation.
\texttt{DF} significantly hampers the model-free robustness with a drop in all transformations.

\subsection{Question Answering (QA)}
\begin{table*}[t]
    \centering 
    \small
    \begin{tabular}{@{}lcccccccc@{}}
        \toprule
        \multirow{2}{*}{\textbf{Method}} & \multirow{2}{*}{\textbf{\makecell{In-Domain \\EM. (\%)}}} & \multicolumn{7}{c}{\textbf{OOD EM. (\%)}} \\
        \cmidrule{3-9}
         & & {DROP} & {RACE} & {BioASQ} &{TBQA} & {R.E.} & {DuoRC} & \textbf{Avg} \\
        \midrule 
        \texttt{{SS}} & 63.76 & 20.09 & 19.29 & 33.91 &28.61 & 62.82 & 32.71 & 32.91 \\ 
        \texttt{{MS}} & \cellcolor{lightgreen}{65.07} & 26.88 & 27.45 & 45.01 & 40.52 & 72.86 & 43.44 & \cellcolor{lightgreen}{\bf 42.69} \\ 
        \texttt{{DA}} & \cellcolor{lightgreen}{63.84} & 19.23 & 19.73& 32.31 & 28.54 &61.97 & 32.31 & \cellcolor{lightred}{32.35}  \\ 
        \texttt{{DB}} & \cellcolor{lightgreen}{64.58} & 20.83 & 19.73 & 34.64 &31.20 & 63.64 & 35.98 & \cellcolor{lightgreen}{34.34} \\
        \texttt{{DF}} & \cellcolor{lightred}{49.56} & 9.25 & 11.72 & 20.94 &19.63 & 45.28 & 21.45 & \cellcolor{lightred}{21.38} \\ 
        
        \bottomrule
    \end{tabular}
    \caption{
    QA Result: Source (IID) accuracy and domain generalization (OOD) on the Question Answering benchmark with NaturalQuestions as source dataset. EM: Exact-Match.
    \textit{See Table~\ref{tab:methods} for method abbreviations.}
    }
    \label{tab:qa_iid_ood}
\end{table*}

\begin{table*}[t]
    \centering 
    \small
    \begin{tabular}{@{}lcccccccc@{}}
        \toprule
        \multirow{2}{*}{\textbf{Method}} & \multirow{2}{*}{\textbf{\makecell{Model Based \\\#Queries}}} & \multicolumn{7}{c}{\textbf{Model Free EM. (\%)}} \\
        \cmidrule{3-9}
         & & {CharSwap} & {EasyData} & {Embedding} &{WordNet} & {CheckList} & {CLARE} & \textbf{Avg} \\
        \midrule 
        \texttt{{SS}} & 19.55 & 60.29 & 52.17 & 61.21 &58.41 & 63.22 & 61.92 & 59.54 \\
        \texttt{{MS}} & \cellcolor{lightgreen}{21.97} & 62.22 & 52.65 & 63.22 &59.84 & 64.42 & 63.55 & \cellcolor{lightgreen}{\textbf{60.98}} \\
         \texttt{{DA}} & \cellcolor{lightgreen}{21.91} & 60.88 & 54.52 & 62.02 & 59.82 & 63.42 & 62.36 & \cellcolor{lightgreen}{60.5} \\
        \texttt{{DB}} & \cellcolor{lightgreen}{20.40} & 61.62 & 53.16 & 62.35 &59.32 & 64.03 & 63.01 & \cellcolor{lightgreen}{60.58} \\
        \texttt{{DF}} & \cellcolor{lightred}{19.19} & 47.97 & 42.48 & 48.55 &47.19& 49.34 & 48.72 & \cellcolor{lightred}{{47.38}} \\
        \bottomrule
    \end{tabular}
    \caption{
    QA Result: Comparison of robustness in terms of model-based evaluation (number of queries needed to fool the model) and model-free (accuracy on adversarial transformations).
    \footnotemark[2]
    \textit{See Table~\ref{tab:methods} for method abbreviations.}
    }
    \label{tab:qa_robustness}
\end{table*}

\begin{table}[t]
    \centering 
    \resizebox{\linewidth}{!}{
    \begin{tabular}{@{}lccccc@{}}
        \toprule
        \multirow{2}{*}{{Method}} & \multirow{2}{*}{{\makecell{In-Domain \\Acc. (\%)}}} & \multicolumn{4}{c}{{OOD Acc. (\%)}} \\
        \cmidrule{3-6}
         & & {MNIST-M} & {SVHN} & {SYNTH} & {Avg}\\
        \midrule 
        \texttt{{SS}}   & 98.40 & 58.09 & 33.85 & 45.94 & 45.96 \\
        \texttt{{MS}}   & \cellcolor{lightgreen}{98.54} & 59.79 & 33.87 & 48.42 & \cellcolor{lightgreen}{47.36} \\ 
        \texttt{{DA}}   & \cellcolor{lightgreen}{99.30} & 67.94 & 42.55 & 48.95 & \cellcolor{lightgreen}{53.15} \\
        \texttt{{DB}}   & \cellcolor{lightgreen}{98.86} & 87.67 & 54.95 & 63.37 & \cellcolor{lightgreen}{68.66} \\
        \texttt{{DF}}   & \cellcolor{lightred}{95.27} & 51.04 & 22.07 & 27.83 & \cellcolor{lightred}{33.65} \\
        \bottomrule
    \end{tabular}
    }
    \caption{
    Source (in-domain) accuracy and domain generalization (OOD accuracy) on the Digits benchmark with MNIST-10k as source dataset.\footnotemark[2]
    }
    \label{tab:mnist_iid_ood}
\end{table}
\begin{figure*}[t]
    \centering
    \includegraphics[width=\linewidth]{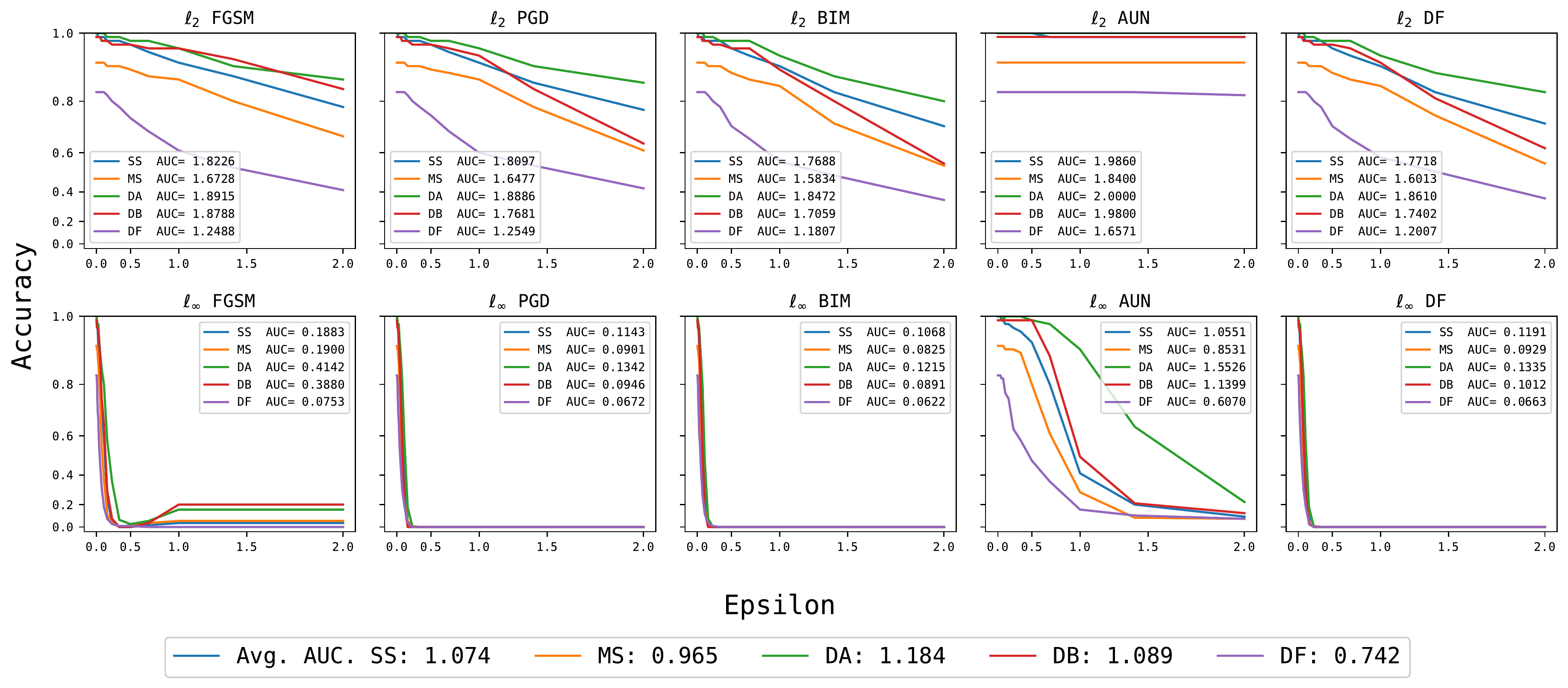}
    \caption{Evaluation of adversarial robustness (using 10 attack methods) for MNIST10k.
    }
    \label{fig:digits_robustness}
\end{figure*}

We focus on extractive QA. 
Given a passage (or ``context'') and a question, the task is to extract the answer span from the passage. 

\paragraph{Methods.}
We use BERT~\cite{devlin2018bert} as the backbone model for each method. 
We use MRQA~\citep{fisch2019mrqa} which is a collection of $12$ publicly available multi-domain QA datasets -- with Natural Questions (NQ)~\citep{kwiatkowski2019natural} as the source dataset.
SQuAD, NewsQA, HotpotQA, SearchQA, and TriviaQA are used as additional datasets for multi-source training. 
Similar to NLI, we use EDA for \texttt{DA} by applying EDA on the question.
We apply the augmentation to all samples in the training set and combine them with the original set to train a \texttt{DA} model.
For model de-biasing (\texttt{DB}), we use Mb-CR approach~\citep{wu2020improving},
where a teacher and bias models are trained \textit{a} priori, and are used for debiasing.

We modify AFLite for our QA task of span prediction, since AFLite was originally designed for classification tasks.
To do so, we first randomly divide the training set into 10 subsets (or folds) $S_{1:10}$.
For $k{\in}\{1,{\dots},10\}$, we pick $S_k$ as the held-out test set, and train models on the rest, and obtain $10$ such models.
At test time, models are used for predicting an answer by only looking at the context (without access to the question) -- this allows us to identify strong spurious correlations in the dataset.
Based on the predictions, samples are sorted on the basis of their F1 score.
A higher F1 score implies that the model is more likely to answer the question without even knowing the question.
We retain $10\%$ samples with the lowest F1 scores -- these represent the task since the model is not likely to predict the correct answer without knowing the question.

\paragraph{Evaluation Protocol.}
We report exact-match (EM) accuracy for MRQA.
To evaluate the generalization performance, we use six OOD development sets from MRQA: DROP, RACE, BioASQ, TextbookQA, RelationExtraction, and DuoRC. 
For robustness, we use the ``Morphues'' attack~\citep{tan2020s} on the question as the model-based evaluation, the attack method is similar to NLI.
Model-free methods are the same as NLI.

\paragraph{Results.} 
Table~\ref{tab:qa_iid_ood} shows the performance of each method in terms of in-domain and out-of-domain accuracy.
We observe that two methods, \texttt{MS} and \texttt{DB}, improve the generalization performance on each out-of-domain dataset and also improve the in-domain performance. 
The improvement of \texttt{MS} is larger than \texttt{DB}. 
\texttt{DA} improves on some out-of-domain datasets but not all, and it also improves the in-domain performance. 
\texttt{DF} dramatically reduces both out-of-domain and in-domain datasets.

Table~\ref{tab:qa_robustness} shows that except for \texttt{DF}, all methods improve over \texttt{SS} for both model-based and model-free robustness evaluation. 
\texttt{MS}, \texttt{DA}, and \texttt{DB} improve the robustness in all transformations of model-free evaluation as well as the model-based evaluation, where \texttt{MS} achieves the best performance in model-based and model-free evaluation. 
\texttt{DF} significantly hampers the model-free robustness with drop in all transformations, meanwhile, the model-based robustness also drops.

\subsection{Image Classification} 

We conduct our experiments on the standard domain generalization benchmark ``Digits'', which is a collection of handwritten digit classification datasets belonging to 10 classes (digits 0--9).
Following standard practice\cite{volpi2018generalizing}, we train models on $10000$ images from MNIST~\citep{lecun1998mnist} as the source, and use SVHN~\citep{netzer2011reading}, SYN and MNIST-M~\citep{ganin2015unsupervised} as the OOD datasets.

\paragraph{Methods.}
We use DigitNet~\cite{volpi2018generalizing} as our backbone image classifier architecture.
Our \texttt{SS} baseline uses MNIST for training; \texttt{MS} uses MNIST and USPS~\citep{denker1988neural}.
For data augmentation we rely on M-ADA~\cite{qiao2020learning} which is a perturbation-based min-max algorithm to create augmented data.
Our debiasing method is RandConv~\cite{xu2020robust} which utilizes a random convolutional layer to generate novel views of each input image, and a KL-divergence based loss function that encourages the classifier to predict consistent predictions for each version of the image.
This leads to the model being debiased on spurious features like background, texture, or color of digits.
We use AFLite as our \texttt{DF} method.

\paragraph{Evaluation Protocol.}
We report IID accuracy on the MNIST test set and generalization as the accuracy on our OOD datasets.
For evaluating adversarial robustness we use Foolbox~\citep{rauber2017foolbox} and use 10 attack methods (both $\ell_2$ and $\ell_\infty$ versions of FGSM, PGD, BIM, AUN, and DeepFool).
Robustness is calculated as the accuracy for 20 values of $\epsilon$ between $[0,2]$, and is plotted as robustness curves for visualization, along with the average values for area under the curve (AUC).

\paragraph{Results.}
Table~\ref{tab:mnist_iid_ood} shows the performance of each method in terms of in-domain and OOD accuracy.
\texttt{MS}, \texttt{DA} and \texttt{DB}, improve the generalization performance on each OOD dataset and also improve the in-domain performance, where \texttt{DB} displays best generalization capacity. 
\texttt{DF} dramatically reduces the OOD performance with significant reduction across all datasets; the in-domain accuracy also decreases. 
Figure~\ref{fig:digits_robustness} shows robustness (accuracy) and area under the curve (AUC) for each plot.
It can be observed that \texttt{DF} is worse than \texttt{SS} for all 10 attack variants.
We observe that \texttt{DA} and \texttt{DB} are better than \texttt{SS}, and the drop for \texttt{DF} is the largest.
\section{Analysis}
\begin{figure}[t]
    \centering
    \includegraphics[width=\linewidth]{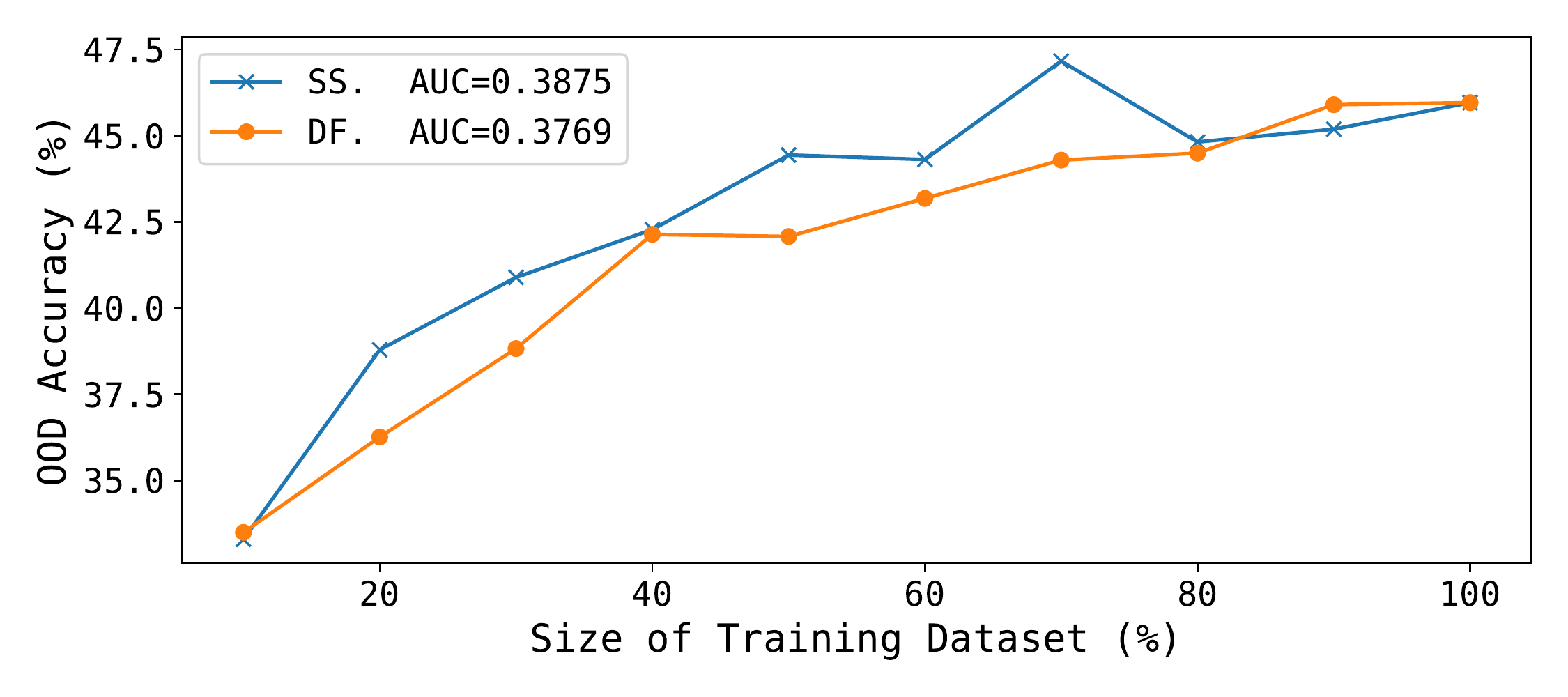}
    \caption{Comparison between SS and DF models trained with different percentages of MNIST10k.
    }
    \label{fig:ood_vs_size}
\end{figure}

\begin{figure}[t]
    \centering
    \includegraphics[width=\linewidth]{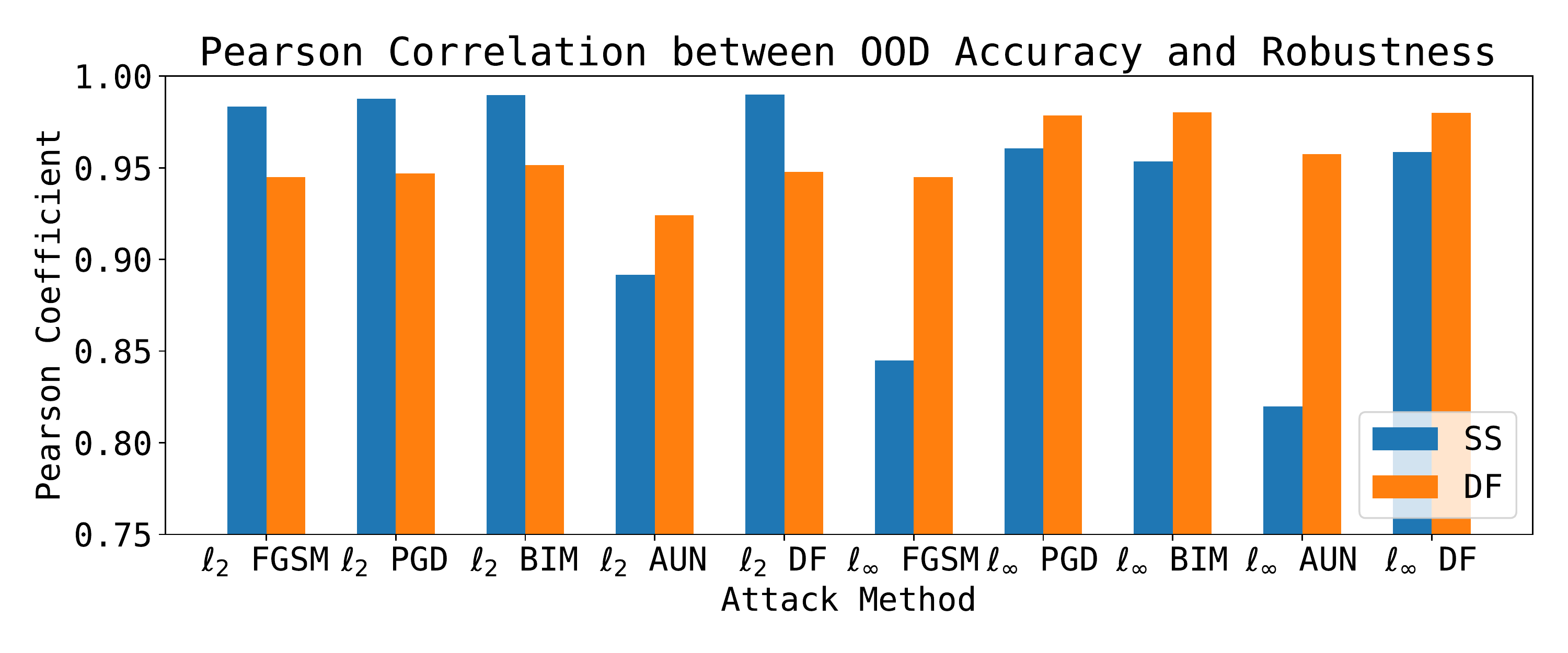}
    \caption{Pearson Correlation between OOD accuracy and robustness for \texttt{SS} and \texttt{DF} models on MNIST10k.}
    \label{fig:correlation}
\end{figure}
Based on the results of three tasks, we have the following observations about the performance of each method compared to the \texttt{SS} baseline:
\begin{itemize}[noitemsep,nosep] 
    \item \texttt{MS} increases OOD accuracy on all three tasks and robustness on two tasks (NLI and QA).
    \item \texttt{DA} increases OOD on two tasks (NLI and IC) and robustness on all three tasks.
    \item \texttt{DB} increases OOD on three tasks and robustness on two tasks (NLI and QA).
    \item \texttt{DF} decreases OOD on two tasks (QA and IC) and robustness on all three tasks.
\end{itemize}

\paragraph{Decrease in NLI in-domain accuracy} is seen for all methods, even though these lead to increase in OOD accuracy.
This suggests that the training dataset (SNLI) has a large shift w.r.t.~OOD datasets.

\paragraph{More data implies more OOD generalization:}
While this trend is observed for both \texttt{MS} and \texttt{DA}, there is one anomaly -- \texttt{DA} for the QA task leads to marginal decrease compared to \texttt{SS} (a difference of $0.56\%$).
This finding is aligned with~\citet{longpre-etal-2019-exploration}, who report no significant effect of data augmentation (back translation) on OOD performance for question answering.
This points to the need for improving data augmentation techniques in QA.
On the other hand, the performance drop due to \texttt{DF} is significantly large for QA ($11.53\%$).

\paragraph{Decrease in MNIST robustness:}
For MNIST, the \texttt{DA} method (M-ADA~\cite{qiao2020learning}) is the best in terms of robustness and also improves OOD accuracy.
M-ADA is an ``adversarial data augmentation'' method, i.e., it uses a min-max objective to find loss-maximizing perturbations and uses these perturbations as augmented data.
It is therefore intuitive that such a method would do well on the adversarial robustness metric (although robustness evaluation was not reported by~\citet{qiao2020learning}).

\paragraph{Marginal Improvement on Robustness:} From the results, it is easy to see that the improvement on OOD is more noticeable than robustness, for example, \texttt{MS} improves OOD performance by ${\sim}10\%$, but improves only by ${\sim}1\%$ under model-free evaluation. 
While this observation is reasonable since each method is designed to improve the generalization, new methods that improve both generalization and robustness should be encouraged.

\subsection{Correlation between Adversarial Robustness and OOD Generalization}
Our experiments reveal the alarming finding that across the board, \texttt{DF} reduces adversarial robustness.
To investigate further, we conduct an analysis on the Digits benchmark and compare \texttt{SS} and \texttt{DF} when trained with equal amounts of data ($\{10\%, 20\%, \dots, 100\%\}$).
Note that for \texttt{SS} the data are sampled randomly, while for \texttt{DF} the data are obtained via AFLite data filtering.
Results are shown in Figure~\ref{fig:ood_vs_size}.
It can be observed that the OOD accuracy increases as the size of the dataset increases, and is greater for \texttt{SS} than \texttt{DF}.
To understand how an increase in OOD accuracy affects robustness, we also compute the robustness values at each size of training data, and compute the Pearson correlation coefficient for each attack method -- positive correlation implies that as OOD accuracy increases, robustness also increases.
Figure~\ref{fig:correlation} shows clear evidence in favor of positive correlation; interestingly, \texttt{SS} has higher correlation for $\ell_2$ attacks, while \texttt{DF}  is higher for $\ell_\infty$ attacks.
The evidence is clear: OOD generalization increases with the size of the dataset and adversarial robustness is positively correlated with OOD generalization.

Our experiments show that the size of the training set directly affects both robustness and generalization.
While removing $90\%$ data increased OOD accuracy in NLI, the effect was the exact opposite for QA and MNIST.
The key idea in domain generalization is that the test distributions are unknown and little information about them is available apart from the fact that there is no task shift.
Without this prior knowledge, deciding whether (or how much) to filter a dataset is a challenging task.

\section{Related Work}
In Section~\ref{sec:categ} we have provided relevant work that falls into one of our five modeling categories.
Here, we discuss additional literature on robustness and generalization and new efforts towards dataset creation, benchmarks, and evaluation.

\paragraph{Generalization Benchmarks.}
\citet{hendrycks-etal-2020-pretrained} have constructed a robustness benchmark for multiple language understanding tasks
by splitting training sets from existing benchmarks according to topics, styles, and vocabulary; this has been subsequently used to study robustness of model rankings~\cite{mishra2021robust}.
Benchmarks have also been constructed to study dataset artifacts and generalization capabilities of models~\citep{Mishra2020OurEM,mishra2020dqi,mishra2020we}.
MRQA~\citep{fisch2019mrqa} is a benchmark for evaluating domain generalization of question answering (reading comprehensive) models. MRQA contains 6 datasets each for training, development, and evaluation.
For image classification, many benchmarks have been proposed to evaluate domain generalization, such as PACS~\cite{li2017deeper}, OfficeHome~\cite{venkateswara2017deep}, Digits~\cite{volpi2018generalizing}, and WILDS~\cite{koh2021wilds} which is a compendium of domain generalization bechmarks for tasks such as image classification, text sentiment and toxicity prediction.

\paragraph{Corruption Robustness.}
\citet{hendrycks2018benchmarking} introduced ImageNet-C and CIFAR-C to test robustness along corruptions such as weather, noise, blur, and digital artifacts, and ImageNet-P which tests robustness against small tilts and changes in brightness.
MNIST-C was introduced by \citet{mu2019mnist} for similar corruptions of handwritten digit images.

\paragraph{Adversarial and Contrastive Sets.}
Generation of adversarial examples~\citep{jia-liang-2017-adversarial, ribeiro-etal-2018-semantically, iyyer-etal-2018-adversarial,alzantot-etal-2018-generating}
and approaches to defend against word substitution~\citep{jia-etal-2019-certified} have been explored. 
Contrastive examples have been introduced as a means for evaluation, for example, manually crafted contrast sets for textual entailment~\citep{gardner-etal-2020-evaluating} or template-based~\citep{mccoy-etal-2019-right, glockner-etal-2018-breaking, naik-etal-2018-stress}.
Model-in-the-loop dataset creation methods have also been proposed for various NLP tasks~\citep{nie-etal-2020-adversarial,arunkumar2020real,kiela-etal-2021-dynabench} and visual question answering~\cite{sheng2021human,li2021adversarial}.

\section{Discussion}
Recently, \citet{miller2021accuracy} have empirically shown linear trends between in-distribution and out-of-distribution performance on multiple image classification tasks, across various model architectures, hyper-parameters, training set size, and duration of training.
They also show that there are certain settings of domain shift under which the linear trend does not hold.
Our work empirically shows that while data filtering may benefit OOD generalization on the NLI benchmark, this does not hold for other tasks such as image classification and question answering.
This suggests that data filtering may benefit generalization in certain types of domain shift, but not on others.
Concurrently, \citet{yi2021improved} have theoretically shown that models robust to input perturbations generalize well on OOD distribution within a Wasserstein radius around the training distribution.
Our empirical observations in this paper in both vision and language domains, agree with the theory of \citet{yi2021improved}.

In this work, we conduct a comprehensive study of methods which are designed for OOD generalization on three tasks:  NLI, QA, and IC.
We evaluate each method on in-domain, OOD, and adversarial robustness.
\footnote{
Code for our experiments will be released at
\url{https://github.com/tejas-gokhale/gen-vs-rob}.}
Our findings suggest that more data typically benefits both OOD and robustness.
Data filtering hurts OOD accuracy on two out of three tasks, and also hurts robustness on all three tasks.
In context of our findings and work by \citet{miller2021accuracy,yi2021improved}, we recommend that methods designed either for robustness or generalization should be evaluated on multiple aspects and not on the single metric that they are optimized for.

\section*{Acknowledgements}
This work was funded in part by DARPA SAIL-ON program (W911NF2020006) and DARPA CHESS program (FA875019C0003). 
The views and opinions of the authors expressed herein do not necessarily state or reflect those of the funding agencies and employers.

\section*{Broader Impact}
One underlying assumption behind using large datasets for training (or pre-training) vision and language models is that larger datasets increase the likelihood of obtaining a diverse set of samples to reduce overfitting.
However, recent studies~\cite{bender2021dangers,stanovsky2019evaluating} serve as cautionary tales when employing uncurated internet data to train large language models, and discuss how large data does not necessarily imply that models will learn the dievrse distribution.
At the same time, the inverse (small data aids diversity) is also not true (as shown by this paper) and comes with its own problems -- for instance, 
Figure~\ref{fig:viz_effect_on_distribution} shows that dataset filtering can lead to much larger changes in the data distribution beyond notions of proportionality and fairness.
As such, the decision on how many and what samples to remove can also introduce its own set of biases.
Data curation is a challenging problem and needs further task-specific study since the concepts of bias and fairness often depend on the task definition and specifications of ideal outcomes.
Insights from this paper could help researchers and practitioners in choosing appropriate approaches for improving generalization and robustness.

\bibliography{tgokhale} 
\bibliographystyle{acl_natbib}




\end{document}